# Associative Emotional Learning in Convolutional Neural Networks

Seowung Leem [1], Andreas Keil [6], Mingzhou Ding [1 *], Ruogu Fang [1,2,3,4,5,*]

[1] J. Crayton Pruitt Family Dept. of Biomedical Engineering, University of Florida, Gainesville, FL 32611, USA

[2] Department of Electrical and Computer Engineering, University of Florida, Gainesville, FL 32611, USA

[3] Department of Computer and Information Science and Engineering, University of Florida, Gainesville, FL. 32611, USA

[4] Department of Radiology, University of Florida, Gainesville, FL. 32611

[5] Center for Cognitive Aging and Memory, University of Florida, Gainesville, FL 32611, USA

[6] Center for the Study of Emotion and Attention, University of Florida, Gainesville, Florida, USA

[*]Corresponding Author:

Ruogu Fang, Ph.D.

J. Crayton Pruitt Family Department of Biomedical Engineering

Herbert Wertheim College of Engineering

University of Florida

PO Box 116131, 1275 Center Drive

Gainesville, FL 32611-6131

Phone: (352) 294-1375

Email: ruogu.fang@bme.ufl.edu

Mingzhou Ding, Ph.D.

J. Crayton Pruitt Family Department of Biomedical Engineering

Herbert Wertheim College of Engineering

University of Florida

Building J285, 1275 Center Drive

Gainesville, FL 32611-6131

Phone: (352) 273-9332

Email: mding@bme.ufl.edu

## Abstract

Associative emotional learning enables organisms to adaptively link pleasant or unpleasant outcomes to the presence of predictive stimuli. Whereas computational models such as the Rescorla-Wagner model have shed light on this important function, the limitations of these models are also known, especially when they are applied to neural data. The advent of deep neural networks has opened another avenue for modeling associative emotional learning. In this work we proposed a deep neural network model of visual valence processing, consisting of a visual module that encodes complex natural scenes and a module that recognizes their emotional significance in terms of valence, a key dimension of emotion, and tested a novel Pavlovian learning paradigm on the model. The results showed that with learning, the model reproduced several observations from human associative learning studies, including association formation and generalization, and that the neural representations of the conditioned and the unconditioned stimuli became increasingly aligned both at the single unit and at the neural population level. Comparison between the model and human experimental data provided further validation of our approach. This study thus suggests that deep neural network models, when combined with appropriate learning algorithms, can be used to model behavioral and neural signatures of associative emotion/valence learning.

***Keywords.*** Computational Neuroscience, Associative Valence Learning, Deep Learning, Pavlovian Conditioning

## 1. Introduction

The formation of associative emotional memories, where discrete, non-emotional events become linked to specific emotional responses such as fear or pleasure, is a fundamental building block of human behavior. Impaired emotional learning is characteristic of mental health problems such as anxiety, post-traumatic stress disorder (PTSD), and substance use disorder (Fullana et al., 2020). Therefore, understanding and modeling the mechanisms of associative emotional learning has both basic and clinical science significance. In human experimental research, Pavlovian (classical) conditioning is one of the key paradigms for probing how individuals form associations between stimuli (Beckers et al., 2023). In this framework, a neutral stimulus, referred to as the conditioned stimulus or CS+, is repeatedly paired with an unconditioned stimulus (US) that carries inherent emotional value, such as aversive electric shock or a rewarding outcome. Over time, the CS+ acquires the capacity to evoke emotional and physiological responses similar to those triggered by the US, measurable through subjective ratings and physiological indices (e.g., skin conductance, heart rate, etc.) (Dunning & Hajcak, 2007; Freund & Keil, 2021; Lonsdorf et al., 2017; Löw et al., 2008).

Classic models of Pavlovian conditioning, such as the Rescorla-Wagner model (Rescorla & Wagner, 1972; Mackintosh, 1975; Pearce & Hall, 1980; Sutton & Barto, 1990), capture many key behavioral observations but do not yield much insight into the underlying neural underpinnings. Neural models based on biophysically grounded neurons can account for fear acquisition, generalization, extinction, and contextual modulations in emotion-processing circuits such as the amygdala (Armony et al., 1995; G. Li et al., 2009; Schmajuk, 2012; Vlachos et al., 2011), but are limited in modeling large-scale changes of neuronal representations through associative learning. More recently, Artificial Neural Networks (ANNs) from the field of deep learning have shown markedly improved predictive power and flexibility (J. Liu et al., 2019, 2021; Shuvaev et al., 2021; Park et al., 2025). These models, while incorporating learning principles into the training process, the training procedures for these models often diverge from the experimental paradigm in empirical Pavlovian conditioning. How to achieve classical conditioning effects in contemporary ANNs under experimental paradigms typically used in empirical neuroscientific settings remains to be addressed.

Neurophysiologically, anterior brain structures such as amygdala, hippocampus, anterior cingulate and medial frontal cortices have traditionally been viewed as the neural substrate of aversive Pavlovian conditioning (Davis, 1997; LeDoux, 1995; Maren & Fanselow, 1996; Rolls, 2023), recent studies have begun to emphasize the crucial role of sensory cortices in emotional processing (Bo et al., 2021) and in the acquisition and maintenance of conditioned emotional responses (Friedl & Keil, 2021a; W. Li & Keil, 2023; You et al., 2021). Convolutional Neural Networks (CNNs), architecturally and computationally inspired by the primate visual system, offer new opportunities to simulate, test, and refine hypotheses about emotional learning, especially concerning the role of the sensory cortex. In particular, representations of CNNs have been shown to replicate stimulus representations in key visual areas such as V1, V4, and the inferior temporal (IT) cortex (Yamins et al., 2014; Khaligh-Razavi & Kriegeskorte, 2014; Güçlü & van Gerven, 2015; Seeliger et al., 2018; Cadena et al., 2019). Moreover, CNNs can be trained to recognize emotions in natural scenes, highlighting the sensory cortex's role in emotional processing (Kragel et al., 2019; P. Liu et al., 2024). Presently, CNN-based models are trained on supervised learning, in which the difference between the model output and the ground truth label is minimized. To what extent these types of models can be trained using Pavlovian learning paradigms and how to evaluate the outcomes of such training remains to be understood.

In this study, we designed an ANN model consisting of a CNN as the visual processing module, which sends the output to an emotion processing module, inspired by the function of the anterior valence-processing structures such as the amygdala and orbitofrontal cortex, to extract the valence of the input. A shortcut connection, motivated by the multi-pathway hypothesis, was further added to link early visual responses from the CNN and the valence-extracting module (Armony et al., 1995; LeDoux, 1996; Rudrauf et al., 2008; Schmajuk, 2012; P. Liu, 2021). The model was first trained to recognize the valence of intrinsically emotional natural images (e.g., those from the International Affective Picture System). A learning paradigm, paralleling the Pavlovian conditioning paradigms used in a large body of empirical psychological and neuroscientific studies (Miskovic & Keil, 2012; Rhodes et al., 2018), was then implemented. Specifically, two Gabor patches, 45 degrees and 135 degrees in orientation, were used as the CS+s and presented together with unpleasant and pleasant images from the International Affective Picture System (IAPS) (Bradley & Lang, 2007) as USs, respectively. During learning, the model was trained to recognize the valence of the compound stimulus (Gabor patch + IAPS picture). The model's capability of associative valence learning was assessed by examining its responses to a CS+ Gabor patch when it was presented alone. Additional questions considered included: (1) neural mechanisms of associative emotional learning, (2) generalization of the learning effects to new contexts, and (3) alignment between the model and human experimental data.

## 2. Materials and Methods

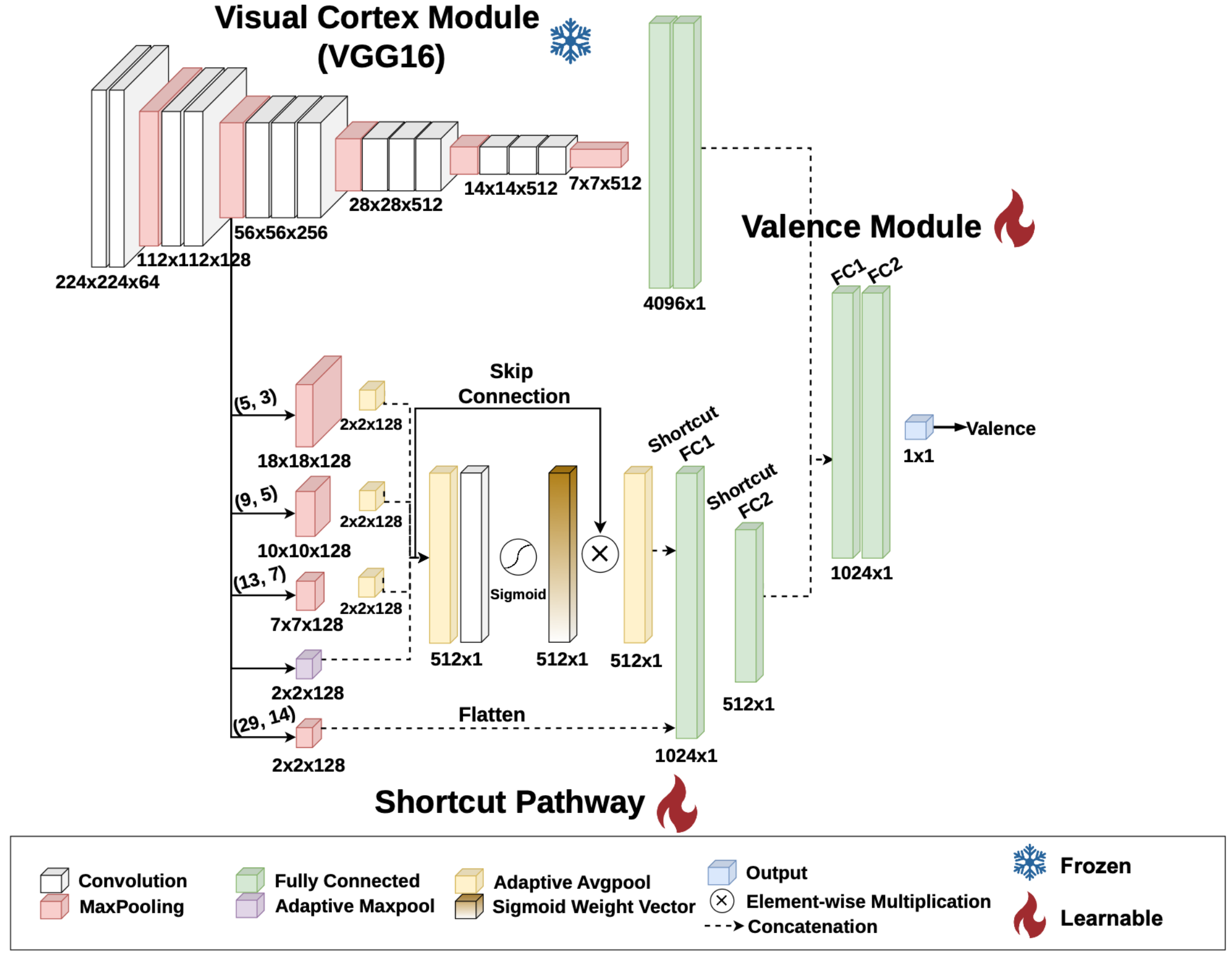

**Fig.1. The Visual-Valence Model for associative emotional learning.** The network model builds on VGG-16, initialized with weights pretrained on the ImageNet object recognition corpus as a model of the visual cortex, to provide a high-capacity visual feature extractor resembling the human ventral visual stream. A parallel Shortcut pathway was added to approximate rapid visual routes from early visual processing to emotion-related circuitry, motivated by evidence of both cortical and subcortical projections connecting early visual processing with higher-order brain structures. Outputs from the Visual Cortex and Shortcut pathways were concatenated and passed to two fully connected layers, the Valence Module, which emulates the anterior emotion processing structures and produces a scalar hedonic valence prediction.

### 2.1. Model Architecture for Conditioning Paradigm

The Visual-Valence Model was designed to receive an input image (224x224) and output the single-digit valence decoding of the input image (See Fig. 1). It consists of 3 major components: a convolutional neural network (CNN) based Visual Cortex module, a Valence Module consisting of two fully connected layers + one-unit output layer, and a Shortcut Pathway module linking the early layers of the Visual Cortex module and the Valence Module. Prior work has shown that CNNs exhibit functional similarities to the human and the nonhuman primate ventral visual stream (Yamins et al., 2014; Kriegeskorte, 2015; Dobs et al., 2022) and that the convolutional window (kernel/filter size) mimics the local receptive fields of visual neurons and multiple convolutional layers model the hierarchical processing of visual features. The particular CNN model considered here is VGG-16 pretrained on the ImageNet dataset (Simonyan & Zisserman, 2015). For the Valence Module, the 2 fully connected layers plus the one-unit output layer was inspired by the three anatomical structures crucial for emotional processing: lateral nucleus (LA) and central nucleus (CE) of the amygdala, and orbitofrontal cortex (OFC), with the LA being the gateway to the amygdala, integrating the information from incoming sensory pathways (Erlich et al., 2012), the CE being the key structure which is in charge of emotional expressions from visual input and generates output to other parts of the brain to coordinate autonomic and behavioral responses appropriate for the goal of the organism (Kalin et al., 2004), and the OFC being a final decision making unit of emotional behavior based on processed sensory and hedonic stimuli (Carmichael & Price, 1995; Hornak et al., 2003; Kringelbach et al., 2003; Rempel-Clower, 2007; Rolls, 2023). In addition to receiving fully processed visual features from the final layer of the Visual Cortex Module, the Valence Module further receives partially processed visual features through the Shortcut Pathway Module, bypassing the

intermediate- and high-level visual processing stages. The Shortcut Pathway was inspired by the multi-pathway hypothesis in emotion processing (LeDoux, 1996; Rudrauf et al., 2008; Catani et al., 2003; Zemmoura et al., 2021) and implemented by (i) extracting the comprehensive features from the early layers of VGG-16 and (ii) processing the computed information using a channel-wise attention mechanism. Pooling layers with different sizes and strides were used to obtain comprehensive features from early visual features from the Visual Cortex module. The channel-wise attention mechanism was implemented by aggregating the features from the pooling layers and processing them using efficient channel attention (Kim & Kwon, 2025; Ramirez-Quintana et al., 2025; Wang et al., 2020). It is worth noting that the attention mechanism considered here refers to a computational mechanism and is not directly related to biological attention.

### 2.2. Model Development for Valence Decoding from Natural Scenes

The model in Figure 1 was first trained to recognize the valence of the visual input. As indicated above, the weights in VGG 16 were held frozen, and the weights in the Shortcut pathway and in the connections between VGG16 and the valence module were trained so that the model acquires the ability to recognize the motivational significance of the visual input, i.e., the valence (on a scale of 1 to 9, with 1 corresponding to extreme displeasure, 9 to extreme pleasure, and 5 to the midpoint of neutral content). The Videoframe dataset (see Section 2.3.1 for image sampling from videoframes) was used by splitting the data into training, validation, and test splits at a split ratio of 8:1:1 (Cowen & Keltner, 2017); see Fig. 2A. The distribution of the valence ratings across the splits was made approximately equal to prevent any emotional bias (Appendix A; Table S1). The learning objective of the model was to decrease the mean squared error between the ground truth valence rating and the predicted valence rating of the image. Additional pictures from the International Affective Picture System (IAPS) were also used. The original valence ratings from Videoframe (Likert scale) and IAPS (Self-Assessment Manikin scale; SAM scale) (Bradley & Lang, 1994) dataset were used as a valence label across this study. Across all training, image augmentations were applied to training and validation sets using a series of methods with corresponding probabilities: rotation (50%), color jittering (20%), Gaussian blurring (20%), and horizontal flipping (50%). The augmentations were not applied to the test set. Using the training and validation data, the model was optimized using the Stochastic Gradient Descent (SGD) algorithm in 50 epochs with a learning rate of 2e-5 and a batch size of 128. Specifically, the momentum of 0.9 and the weight decay of 1e-4 were applied in the SGD. In addition, the learning rate decayed with a multiplicative factor of 0.1 was applied for every 20 epochs. During the training, the best model was collected, defined as the one achieving the smallest loss on the validation set.

The unconditioned stimuli used in the conditioning paradigm were images from the IAPS (see below). Therefore, the model trained on the Videoframe dataset was further finetuned with the IAPS dataset to minimize the effect of stimulus variations between the datasets. 10% of the IAPS data were used for training, 10% for validation, and the remaining 80% for testing. In this way, the vast majority of samples were reserved to evaluate the finetuned model. The model was optimized with a learning rate of 2e-4 and a batch size of 10, with 100 epochs. Given that in the conditioning paradigm, the CS+ and the US are presented in the second and the fourth quadrants, respectively, additional tuning was performed by reducing the field of view by placing the same IAPS dataset on 4th quadrant. The tuning was performed with 100 epochs, with a learning rate of 1e-5 and a batch size of 16. The final model, after finetuning on quadrants, was used in each conditioning scenario.

To evaluate the model's performance on emotional decoding, the Pearson correlation coefficient between the model's predicted valence and human-labeled valence of input stimuli was acquired to provide the behavioral level performance of the model. In each training step (after training with Videoframe and finetuning with IAPS), the model was evaluated with the test split corresponding to the source training data.

**2.3. Affective Datasets**

More details of the two affective image datasets considered here: (1) the Videoframe dataset and (2) the IAPS dataset are provided below.

*2.3.1 Cowen & Keltner Videoframe Dataset*

The Videoframe dataset (Cowen & Keltner, 2017) consists of 2,185 short videos designed to evoke emotional experiences. Each video has associated normative valence and arousal ratings, collected from 853 (403 female) English-speaking US participants using Amazon Mechanical Turk. The ratings were based on a nine-point Likert scale with 5 as neutral, 9 as most pleasant, and 1 as most unpleasant. In this study, the Videoframe dataset was used to develop the model's emotion prediction capability. Because the model used in this study only accepted static pictures, to convert the videos into a set of static images, for a given video, every 10th frame was extracted and assigned the video's valence. The dataset had a total of 21,876 affective images.

*2.3.2 International Affective Picture System.*

The International Affective Picture System (IAPS) dataset consists of 1182 images, and each image has a normative valence rating collected from human observers (Bradley & Lang, 2007). The ratings of the image were based on the SAM scale, from 1 as most unpleasant to 9 as most pleasant. Unconditioned stimuli (appetitive/pleasant and aversive/unpleasant) for the conditioning experiment were taken from this dataset. Specifically, 22 images with the most aversive valence (valence = 8.09±0.12), and 22 images with the most appetitive valence (valence = 1.55±0.10) were used as unconditioned stimuli. The purpose for using these extreme images was to achieve the most robust conditioning effects. The rest of the 1,138 images were used to fine-tune the model used in this study.

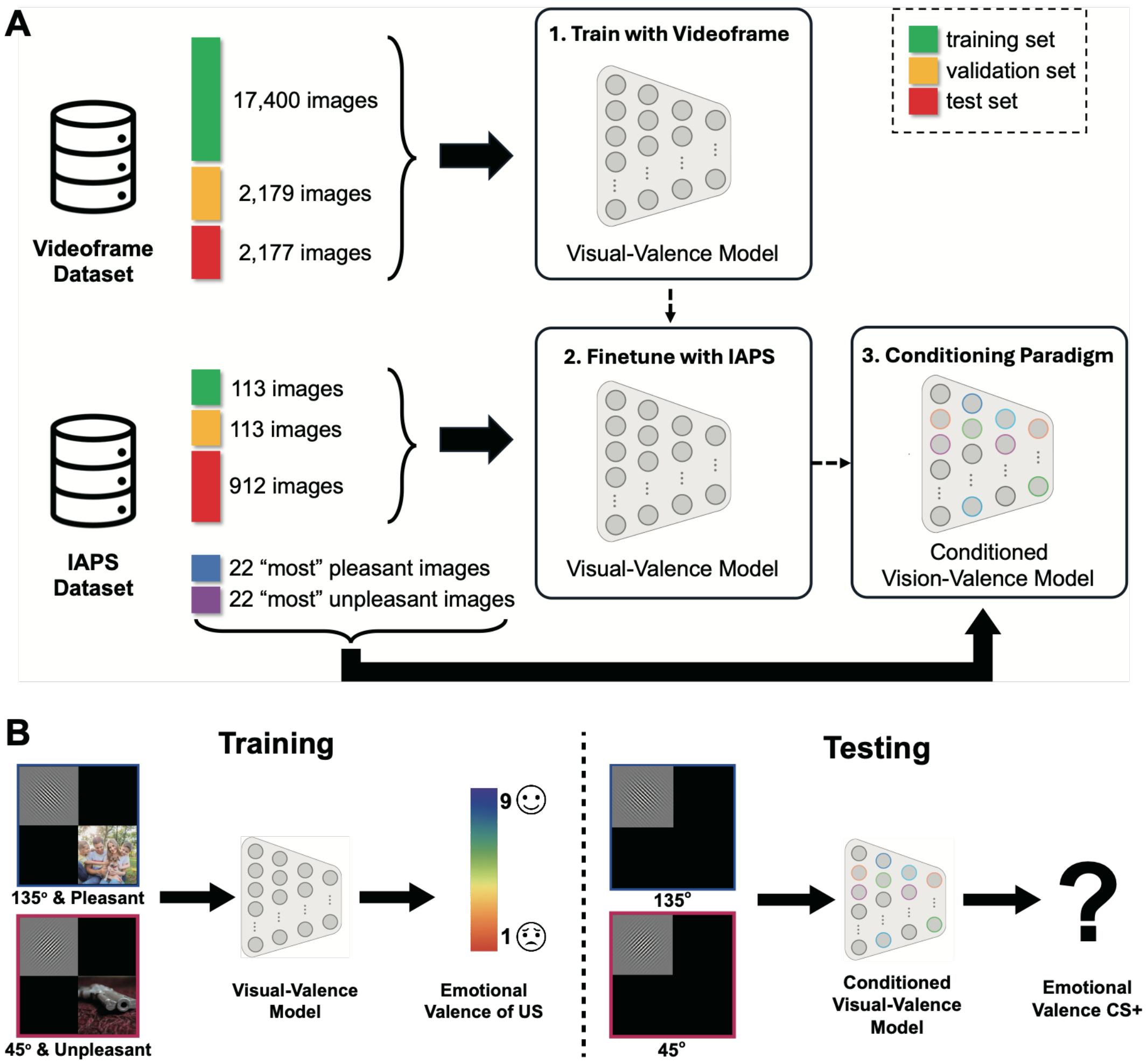

**Fig. 2. Model Training and Pavlovian conditioning paradigm.** (A) Training and fine-tuning of the Visual-Valence model for achieving valence decoding. The model was first trained on the Videoframe dataset, which was split into training, validation, and test sets with 8:1:1 ratio. The trained model was finetuned with the IAPS dataset, with 10% of the images for training, 10% for validation, and 80% for testing. From the IAPS dataset, the 22 most pleasant and 22 most unpleasant images based on the valence were selected as unconditioned stimuli (US) for the conditioning paradigm. The final model after quadrant-specific finetuning was used in the following conditioning experiment. (B) Schematic overview of the training and test of the conditioning paradigm. Each trial presents a compound stimulus consisting of a Gabor patch (an oriented grating) as the conditioned stimulus (CS+) and an affective natural image from IAPS as an US. The 45° Gabor patch was always paired with unpleasant images, whereas the 135° Gabor patch was always paired with pleasant images. The unconditional stimulus (i.e., an IAPS picture) was placed in the 4th quadrant. The conditional stimulus (CS+) was presented in the 2nd quadrant. These locations were fixed throughout the learning process. The task of the model was to predict the valence rating of the US for the compound stimulus. Bottom (testing): During the testing phase, Gabor patches of different orientations, spatial frequencies, and contrasts, as well as appearing in different quadrants, were presented, and the valence prediction from the model was obtained. (Note: The US images shown in this figure are shown for illustrative purposes only and were not part of the experimental stimulus set. The "family portrait" image was obtained from StockSnap (https://stocksnap.io/photo/family-portrait-6G9JL1XVYS), and the "gun" image was obtained from Rawpixel (https://www.rawpixel.com/image/5964101); both are released under the CC0 1.0 Universal license.)

### 2.4. Conditioning Paradigm for Deep Neural Networks

After the model in Figure 1 was trained and finetuned to acquire the ability to recognize the valence of the visual input, the following learning paradigm was then implemented to test associative emotional learning.

*2.4.1 Overview*

Fig. 2(B) shows the Pavlovian conditioning paradigm. Gabor patch gratings were used as CS+s. In particular, 45° and 135° Gabors were CS+s, associated with unpleasant and pleasant USs, respectively. The pairings of CS+s and US were performed for 100 epochs, with randomized pairings in every epoch. To understand how generalizable our paradigm is, we also tested frequency conditioning and contrast conditioning; for frequency conditioning, 0.1 and 0.3 cycles per pixel were CS+s associated with unpleasant and pleasant USs, respectively, and for contrast conditioning, 20% and 80% contrasts were CS+s associated with unpleasant and pleasant USs, respectively, and included the results in the Supplementary Information (Fig. S3). As the orientation conditioning depicted in Figure 2 is the most commonly used paradigm in empirical settings, the focus of this paper was primarily placed on orientation conditioning. Each conditioning scenario was repeated 10 independent times with different random seeds to account for the variability of valence prediction.

*2.4.2 Paradigm (Training & Testing)*

The association of affective stimuli to neutral cues was implemented by presenting natural affective images (unconditional stimulus; US) and Gabor patches (conditional stimulus; CS+) simultaneously in two different quadrants: The US was always placed in the 4$^{th}$ quadrant, and CS+ was always located in the 2$^{nd}$ quadrant during both the learning and testing stages. During the learning phase, the model's task was to predict the valence of the US for the CS+-US compound stimulus. The optimization method of the model was the same as the method used in the training & fine-tuning step, except that the learning rate was chosen to be 1e-4 and batch size 10. In the testing phase, only the CS+ (i.e., Gabor patches) was shown to the model, placed in the 2$^{nd}$ quadrant. The valence predictions by the model were obtained to assess whether, after associative emotional learning, the neutral stimuli acquired the ability to evoke emotional responses. For generalization testing, the valence ratings of 13 different orientations (from 0° to 180° with 15° differences) were acquired, which included the two CS+s and 11 other Gabor gratings never seen by the model during the learning stage. In addition, the valence ratings of CS+s placed in different quadrants were obtained to test generalization across spatial locations.

## 2.5. Analysis of Neural Activation and Representation in the Model

### *2.5.1 Individual Neuron Level Correlation Analysis*

To characterize how associative learning reshapes the neural activation to stimuli, unit-level responses to the CS+ and US were analyzed across the artificial neurons in the four trainable network layers (Shortcut FC1, Shortcut FC2, FC1, FC2) pre- and post-conditioning (see Fig. 1.). Co-selectivity was indexed by the Jaccard index, measuring the degree of overlap between units activated by the US and by the CS+. Within each layer, Pearson's correlation was then computed between the CS+ and US evoked response neurons, reported both for all units and separately, for the subset co-active to both stimuli. The significance of correlation was obtained by a two-sided t-test under Student's t distribution. This analysis was to examine whether the neurons responsive to US also became responsive to CS+ through associative learning.

### *2.5.2. Population Level Decoding Analysis via Support Vector Machine*

Extensive evidence has shown that population-level neural activity patterns encode information not possible at the single neuron level. The decoding analysis framework was set to read out (1) emotion (pleasant vs unpleasant) and (2) orientation (45° vs 135°) from network population features. The support vector machines (SVM) with a Gaussian kernel ($\gamma$=1/number of features) were used as a decoder to perform the analysis. The features were neural activity across a layer standardized within a preprocessing pipeline using the mean and variance computed on the training split only to prevent data leakage. Unless noted, decoding was performed separately on features taken from each of four fully connected layers (Shortcut FC1, Shortcut FC2, FC1, FC2). Each decoding analysis was performed 100 times, and statistical significance was computed by the one-sample t-test.

To compare the representations of the US and CS+ after associative emotional learning, the post-conditioning model was used as a feature extractor. Using the feature extractor, US (n=44; 22 unpleasant and 22 pleasant) and CS+ (n=2; 45° and 135°) representations were obtained from 4 layers. Then, the US representation was split into training/test sets into half (22 images per set) with a stratified split to maintain the same number of examples for each emotion. The decoder was fit with a training set. The fitted decoders were tested with US representations in the test set and CS+ representations. For CS+, 45° was considered unpleasant, and 135° was considered pleasant in this decoding analysis. To test whether the encoding of the physical stimulus property (e.g., orientation) associated with the CS+ (grating orientation) was preserved after associative learning, an orientation decoder was fitted independently of outcome valence. Two orientations (45° and 135°) were augmented with spatial frequencies (0.02~0.4 cycles per pixel, 0.02 interval) and spatial contrast (0~100%, with 10% interval), yielding 220 images per orientation (440 in total). For each image, feature vectors were extracted from the four fully connected layers pre- and post-conditioning. A decoder fit to pre-conditioning features was then evaluated on features from both the pre-conditioning and post-conditioning networks, providing a measure of orientation encoding and its persistence following learning.

## 2. Results

### 3.1. The Visual-Valence Model

The Visual-Valence Model used in this study consisted of a Visual Cortex module modeled by VGG-16 pretrained on ImageNet (Deng et al., 2009), a Valence module modeled by two fully connected layers plus a one-unit output layer, and a Shortcut Pathway connecting the early layers of the Visual Cortex module to the Valence module. The Valence module thus receives both low- and high-level visual representations of the input for mapping into emotional representations (Fig. 1). To endow the Visual-Valence model with the ability to assess the emotional significance of visual input, we first trained the model to predict the hedonic valence of natural scenes on a scale from 1 to 9, with 1 corresponding to extreme displeasure, 9 to extreme pleasure, and 5 to the midpoint of neutral content (Fig. S1). The decoding performance model for each test split (Videoframe=0.4007; IAPS=0.4395; result of train and validation split available in Appendix A). These results indicate the model had acquired the ability to reliably capture the hedonic valence of natural visual input.

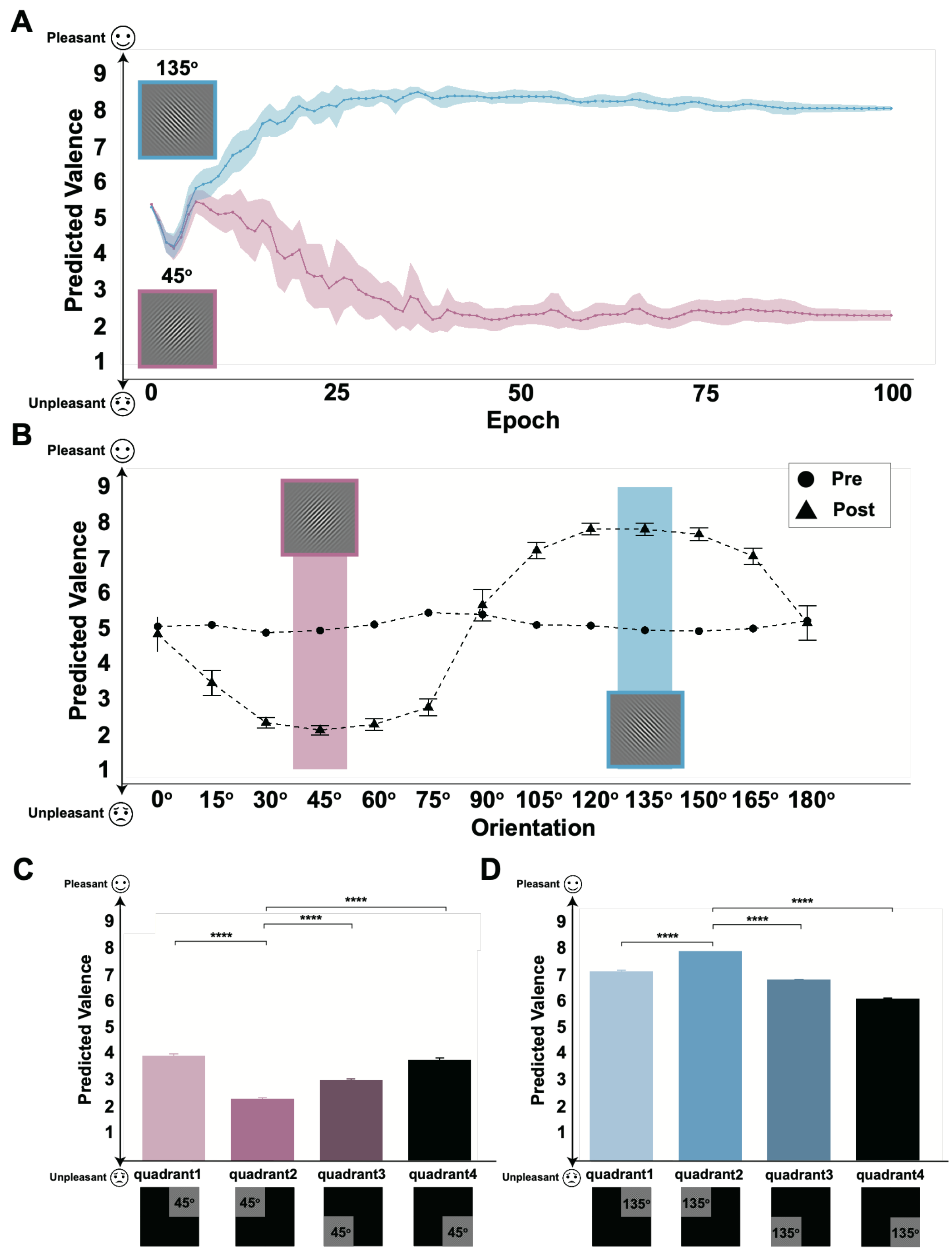

**Fig. 3. Model performance during and after associative emotional learning.** (A) Model predicted valence of the conditioned stimuli (CS+) as a function of training epochs. The result of 10 independent associative learning experiments with different random seeds is shown in the figure, with the mean value displayed as a solid line and the standard error represented by shaded regions.

The two different conditioned stimuli (CS+) used in the associative learning process are displayed in different colors (violet for unpleasant, cyan for pleasant). Through associative learning, the model was able to differentiate the opposite emotional significance of the two CS+ gratings, consistent with findings from differential Pavlovian conditioning. (B) The predicted valence across 13 different orientations, including 11 orientations not used during training, demonstrates generalization by the trained model. It is noteworthy that before associative learning, the model assigned neutral valence ratings to all 13 stimuli (~5). Bar plot shows the valence prediction result of (C) 45°orientation and (D) 135° orientation placed in different quadrants after associative learning, demonstrating generalization across spatial locations. Here, the result was obtained by performing 10 different iterations. Error bars indicate the standard error of 10 different iterations, and statistical significance was calculated using a corrected t-test. N=10, ****P<0.0001, Bonferroni corrected. (Note: The y-axis in the figures is labeled with symbolic face markers to show the meaning of valence intuitively.)

### 3.2. Associative Emotional Learning in the Visual-Valence Model

The Pavlovian conditioning paradigm is shown in Fig. 2B. There were two phases. In the training phase, the US and CS+ were shown simultaneously in different quadrants as a compound stimulus to the model. For the US, we used 22 unpleasant images (valence mean ± std; 1.31±0.10) for aversive conditioning and 22 pleasant images (valence mean ± std; 8.09±0.12) for appetitive conditioning, where all images were taken from IAPS. For the CS+, we used gratings oriented at 45 and 135 degrees relative to the vertical axis with 45 degrees always paired with the unpleasant images and 135 degrees with the pleasant images. On each trial, the CS+ and the US were presented in the 2nd quadrant and the 4th quadrant, respectively, and the network was trained to predict the valence of the US. In the testing phase, CS+ was presented alone in the 2nd quadrant, and the predicted valence by the model was recorded and analyzed to examine whether the CS+ had acquired appetitive or aversive properties through associative learning and conditioning (see Section 2.5 for details).

Fig. 3A shows the predicted valence of the two CS+ as associative learning progressed; here, the learning progression was indexed by the epoch number. At the start of the learning process (epoch 0), the valence of both CS+ was predicted by the model to be in the neutral range (for 45° and 135°orientations, the model predicted valence =5.37 & 5.48, respectively). While this result was somewhat expected because, at Epoch 0, CS+ has not been associated with emotional outcomes, it is nevertheless noteworthy as the model at this point has never been exposed to Gabor patches (i.e., there were no Gabor patches in the ImageNet database). After aversive and appetitive associative learning, the model predicted emotion significance of the 45° CS+ transitioned from neutral to unpleasant (valence=2.31±0.15) and the 135° CS+ from neutral to pleasant (valence=8.01±0.06). A similar pattern was found when CS+ was parametrized along the dimension of spatial frequency, but not when it was parametrized along the dimension of spatial contrast (Fig. S3; Appendix B.2). The association was not achieved when the valence of the US was randomized (Fig. S4A; Appendix B.3) and the learning was present with CS-US pairing was reversed (Fig. S4B; Appendix B.3).

Having demonstrated that our model is capable of associative emotional learning, we next asked how the model responds to novel stimuli that bear resemblance to the CS+ but have never been shown to the model during learning. More specifically, we tested whether the model is capable of generalization, a well-established property in experimental emotional conditioning for both mammals and humans (Dunsmoor et al., 2009; Hull, 1943; Lissek et al., 2008; Pavlov (1927), 2010). Two tests were performed.

First, we tested the generalizability of our model to Gabor patches of different orientations (Fig. 3B). Specifically, Gabor patches of 13 different orientations (spanning 0° to 180° at 15° intervals) were presented to the Visual-Valence model pre- and post-conditioning. As expected, the valence for all 13 gratings was rated to be neutral (mean±s.d.; 5.08±1.62) before learning (pre-conditioning), implying that the deep neural network viewed the artificial stimuli as neutral despite the fact that it was never exposed to such stimuli during its training process. After conditioning (post-conditioning), the predicted valence across the 13 Gabor patches exhibited the classic generalization pattern, in which aversiveness and appetitiveness responses evoked by the stimuli in the model gradually diminished as the orientation moved away from 45° & 135°, respectively; in other words, the effect of associative learning was stronger when stimuli bore greater similarity to a given CS+. The monotonic generalization was observed in patterns across different contrast conditions in the orientation conditioned model (Fig. S5; Appendix B.4).

Second, we analyzed the model's spatial generalizability by presenting the CS+ in all 4 quadrants of the input image, including those where CS+ had never appeared during the learning phase, i.e., first, third, and fourth quadrants. The bar plots in Fig. 3C and Fig. 3D show the valence prediction result of two CS+ (45° and 135°). The emotional learning effects were most significant when CS+ was placed in the 2nd quadrant (mean±s.d.; 45°=2.31±0.15, 135°= 8.01±0.06), as expected, because the 2nd quadrant was where the CS+ appeared in the learning process, and became weaker but still significantly compared to the neutral valence in pre-associative learning when it was placed in the other three quadrants (1st quadrant;

45°=3.95±0.26 $t(9)=28.6$, $p<0.0001$, 135°= 7.10±0.19 $t(9)=19.4$, $p<0.0001$, 3rd quadrant; 45°=3.03±0.19 $t(9)=15.4$, $p<0.0001$, 135°= 6.79±0.12 $t(9)=32.6$, $p<0.0001$, 4th quadrant; 45°=3.80±0.23 $t(9)=19.8$, $p<0.0001$, 135°= 6.07±0.12 $t(9)=57.8$, $p<0.0001$). These results demonstrate that emotional learning effects generalize across spatial locations. This is similar to what is observed in human experiments (Friedl & Keil, 2021b).

To show that the conditionings effects were not specific to the particular type of CS+ used in the paradigm, using bar-shaped stimuli as a CS+, we found comparable conditioning outcomes (Fig. S6; Appendix B.5). In addition, to determine whether the conditioning effects reflect actual effects of learning rather than effects driven by low-level visual properties of the CS+ (e.g., orientation or contrast), we conducted a partial regression analysis controlling for preconditioning responses and stimulus features (Appendix C). The results indicated that the network's predictions were driven by higher-level semantic structures, which is valence from the US in our study, rather than memorization of pixel-level features associated with the CS+ stimulus.

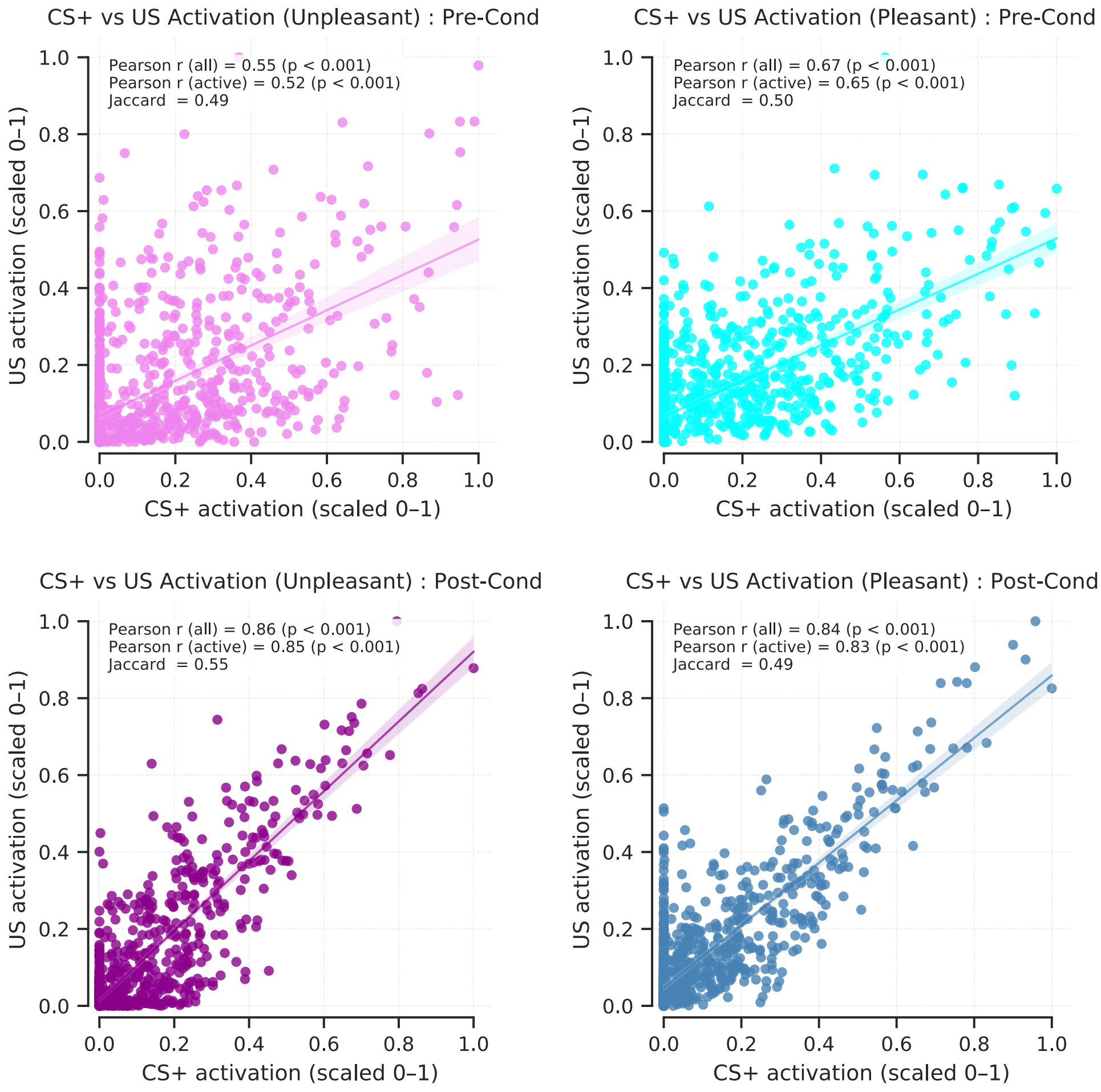

**Fig. 4. Unit-level responses to CS+ and US.** The violet (unpleasant) and cyan (pleasant) colors indicate the single neuron activation evoked by the unconditioned stimulus and the conditioned stimulus in layer (FC2) pre- (top row) and post- (bottom row) conditioning. The points were min-max scaled to 0 and 1, and the solid line represents a regression fit with a 95% confidence interval (shade). The metrics are (1) Pearson r (all): correlation computed across all neuron activation, (2) Pearson r (active): correlation computed across neurons active for both US and CS+, (3) Jaccard: co-activation index for neurons active for both CS+ and US. Statistical significance was calculated by the two-sided t-test. Metrics are derived from the original activations; scaling is for visualization purposes only.

### 3.3. Neural Mechanisms of Associative Emotional Learning

It is expected that the neural responses to US and CS+ would become more similar as learning progressed. To determine whether our model conforms to this pattern, we quantified the impact of associative learning on the representation of the stimuli. First, unit-level responses across all trainable layers were analyzed. Specifically for each layer, the unit activations in response to CS+ (45° and 135°) and US (unpleasant and pleasant) were extracted. The activations evoked by US stimuli were averaged to yield a single US activation. Then, Pearson correlation between CS+ and US across all neurons within each layer before and after associative learning was computed. The correlation measures the degree of the learned representation of CS+ converging toward US, providing an interpretable change induced by the associative learning. Before associative learning, the last layer immediately preceding the valence output (FC2) exhibited the greatest co-activation (Jaccard index; unpleasant (Pre-Cond) =0.49, pleasant (Pre-Cond) =0.50, unpleasant (Pre-Cond) =0.55, pleasant (Pre-Cond) =0.49; Fig. 4) relative to earlier layers (Fig. S7; Appendix B.6). These findings indicate that learning-related changes should be most apparent in the last layer. Following associative learning, unit-wise responses to CS+ and US became more similar, and correlations increased by 0.33 (unpleasant) and 0.18 (pleasant). This convergence strengthened with layer depth (Shortcut FC1: $\Delta r = 0, 0$; Shortcut FC2: $\Delta r = -0.06, 0.03$; FC1: $\Delta r = 0.15, 0.08$). Consistent with the partial regression analysis of model behavior, we performed a similar analysis on neuronal activations while controlling for preconditioning responses and low-level visual properties of the Gabor patch stimuli (Appendix C). The results indicated that the representational structure of the CS+ was modulated by associative learning. Thus, at the single neuron level, associative learning increased the alignment of the activation evoked by the CS+ and the US.

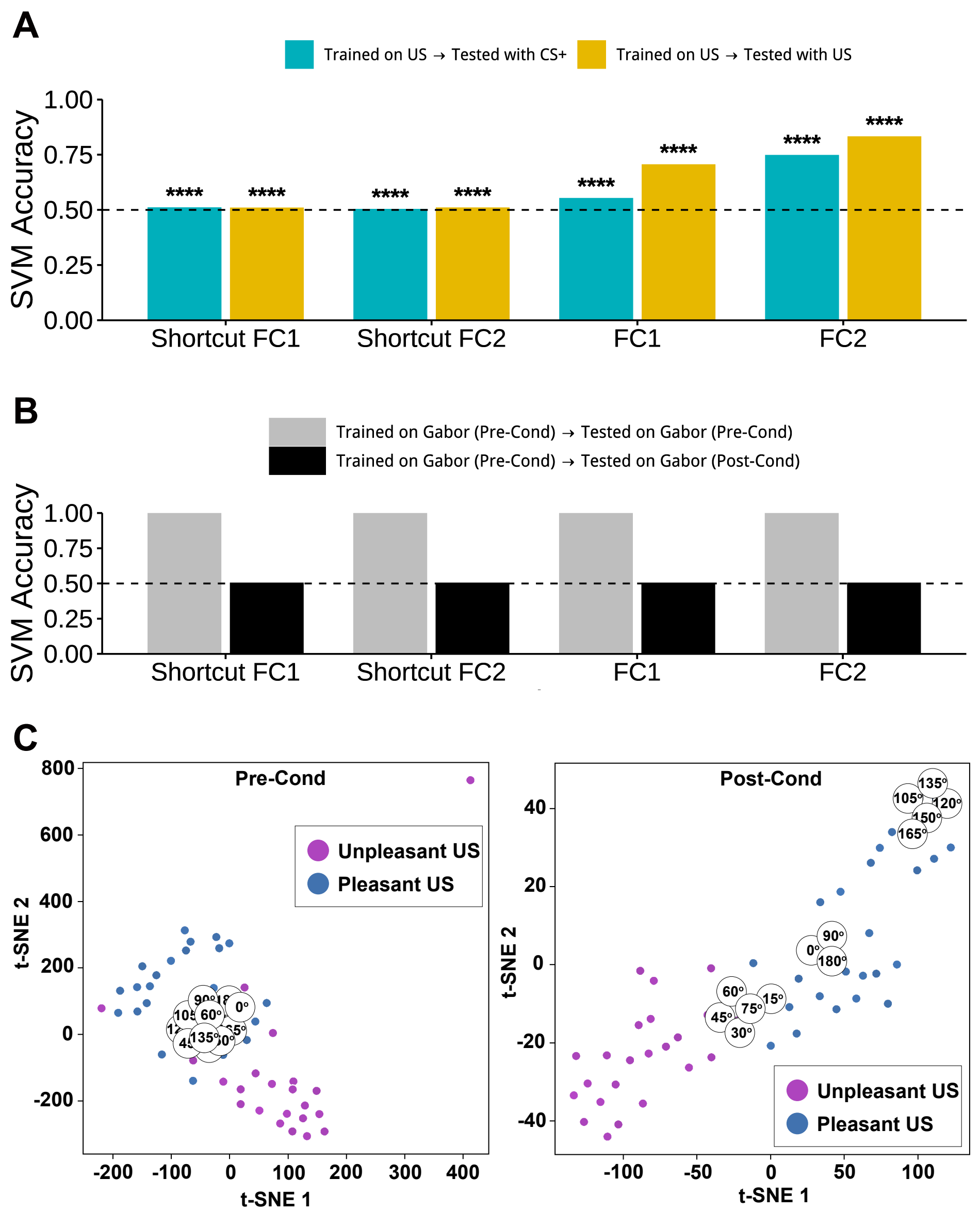

**Fig. 5. Population-level representations of CS+ and US.** (A) Classification accuracy of the support vector machine (SVM) emotion decoder fitted on half of the US feature representations and tested on the remaining US (yellow) and CS+ (teal) representations across 4 learnable layers.

The dashed line marks the 50% chance level. 100 different iterations were performed to derive statistical significance (****P<0.0001, one-sample t test). (B) Classification accuracy of the SVM orientation decoder fitted on the orientation representation of pre-conditioning representations of 45 and 135 degree gratings. The dashed line marks the 50% chance level reference, and the result of 100 different iterations is shown. The representation of orientation gratings from the model before (Pre-Cond) and after conditioning (Post-Cond) is tested. All results were statistically significant (P<0.0001, one-sample t test). (C) t-distributed stochastic neighborhood embedding (t-SNE) results show how the embedding of CS+ and US is modified by appetitive and aversive conditioning. The population-level representations of CS+ and US were obtained from the FC2 layer of the model. The colors of the data points indicate the emotion class of the US. The embeddings of 13 different orientations were labeled inside each data point. The embeddings of pre- and post-conditioning features were performed separately; therefore, the distance between the data points in the figure is not comparable.

To further demonstrate the population-level or pattern-level characteristic of the neural process under associative learning, we trained a series of support vector machine (SVM) decoders. The first set of decoders was trained on US images to discriminate unpleasant versus pleasant images, using layer-wise activation patterns as feature vectors (Fig. 5A). This process led to 4 different SVM classifiers: each SVM classifier for each of the four fully connected layers (see Section 2.5.2 for details). The objective of the SVM was to classify the two CS+s and the remaining US into two emotional classes. When evaluated on US images, decoding accuracy increased with depth (mean ± s.d.; chance level=0.5000, Shortcut pathway FC1 = 0.5068±0.1042, Shortcut pathway FC2 = 0.5073±0.0861, FC1=0.7027±0.0862,

FC2=0.8295±0.0836). This indicates that the affective information is concentrated in the deeper layers. Applying the same decoders to CS+ yielded a parallel depth-dependent pattern (Shortcut pathway FC1 = 0.5075±0.0321, Shortcut pathway FC2 = 0.5008±0.0050, FC1=0.5500±0.2072, FC2=0.7450±0.2512), suggesting that at the neural population level, the CS+ representation became more similar to US representations after associative learning.

A second set of decoders was trained to discriminate the orientations of the two CS+s (45° and 135°) using layer-wise activation patterns as features to probe representational remapping (Fig. 5B). One decoder was fit per learnable fully connected layer with pre-conditioning activations and used to classify pre- and post-conditioning activations of two orientations. On pre-conditioning activations, all decoders performed at ceiling (chance level = 0.5000; accuracy = 1.0000 across layers), confirming robust orientation encoding. When evaluated on post-conditioning activations, performance collapsed to chance level (accuracy = 0.5000 across layers), indicating that the original orientation information in the CS+ representation was effectively overwritten by a new emotional code driven by associative learning.

In Fig. 5C, the t-distributed stochastic neighborhood embedding (t-SNE) offers an additional intuitive visualization of the disposition of different Gabor patches and the unconditioned stimuli within the feature space of the last fully connected layer. Pre-conditioning, Gabors of different orientations were clustered in the middle between the unpleasant and pleasant images, whereas in post-conditioning, Gabors of different orientations were divided into three groups clustered together with the emotional images. Thus, the high classification accuracy of the SVM trained with the features from the last fully connected layer is visually explained by the t-SNE embeddings. Furthermore, embeddings post-conditioning suggests the learning process induced convergence between feature representations of the conditioned and unconditioned stimuli, rather than merely aligning the orientation embeddings with unchanged emotional embeddings.

## 3. Discussion

In the present study, we set out to examine the extent to which deep neural networks can be used as a model for associative emotional learning. We first developed a Visual-Valence model, a deep neural network consisting of a visual cortex module, an emotion evaluation module and a shortcut pathway connecting early visual processing with the emotion module, and trained the model to recognize the hedonic valence (pleasure versus displeasure) associated with intrinsically natural scenes. We then introduced a novel learning paradigm, in which neutral stimuli (Gabor patches) differing along a fundamental feature dimension, such as orientation, were paired with pleasant or unpleasant pictures, and the network was asked to assign the same valence as the emotional pictures to the compound stimuli. The valence ratings were then obtained for Gabor patches presented without the emotional pictures. We found that the model learned, within dozens of paired presentations, to assign pleasant or unpleasant valence to previously neutral visual stimuli that were consistently paired with pleasant or unpleasant outcomes, and that conditioned emotional evaluations generalized to similar stimuli in a fashion that resembled generalization in humans. Generalization was most pronounced for orientation conditioning, compared to contrast or spatial frequency conditioning, paralleling what has been observed with human participants (Friedl & Keil, 2021a). Unit-level and population-level neural representation analysis revealed that learning prompted by associative emotional conditioning mainly took place in deeper structures of the DNN model. Interestingly, as shown in Supplementary Materials (give details), as learning progressed, the neural activation patterns in these structures increasingly resembled those of human electrophysiological recordings.

In the field of affective neuroscience, diverse neurophysiological models of emotional stimulus processing have been proposed based on brain imaging research (Pessoa & Adolphs, 2010; LeDoux, 1996). Our model's architecture is based on having a sequential connection from visual input to anterior brain structures modulated by emotion, such as the amygdala and orbitofrontal cortex (OFC), and a shortcut pathway from early visual cortex to these structures, to simulate faster but less refined visual information (Kawasaki et al., 2001; Streit et al., 2003). Although the biological circuitry (e.g., cortico-visual pathway as well as subcortical visual-emotion system) is far more complex than our simplified architecture, our experiments show that the model, when combined with a novel associative emotional learning paradigm, was able to produce the well-known conditioning effects where previously neutral stimuli (e.g., Gabor patches) were recognized as emotionally significant. In addition, neural mechanism analysis supports the core hypothesis of classical conditioning, which is that learning is done by increasing the representation similarity between conditioned and unconditioned stimuli. In particular, our model performs the learning by adjusting the hierarchical feature representations so that the conditioned stimulus comes to resemble the unconditioned stimulus. More broadly, the results underscore the utility of CNNs for computational studies of associative learning, owing to the fact that CNN-based architectures provide rich, layered sensory encodings of conditioned cues, enabling models to process complex visual inputs in a manner analogous to mammalian perception. This capacity makes them well-suited for probing how stimulus features are transformed during learning and for building models that bridge perception and affective processing. Additionally, in Supplementary Materials, it is shown the Visual-Valence model without the shortcut pathway was not able to demonstrate emotional associative learning (Fig. S2; Appendix B.1), lending support to the multi-pathway hypothesis of emotional processing (LeDoux, 1996; Rudrauf et al., 2008; Catani et al., 2003; Zemmoura et al., 2021).

In the context of recent neurocomputational theories, the amygdala is thought to play a role in affective processing and automatic responses, and signals from the orbitofrontal cortex encode the reward value of a given stimulus (e.g., Rolls, 2023). In this framework, appetitive and aversive dispositions arise from the brain systems that represent reward and punishment, and stimulus-reinforcer associations are learned through OFC and other prefrontal regions. In our Visual-Valence model, the associative learning between sensory stimuli and valence signals was implemented similarly. Moreover, conditioned valence representations mainly occurred in deeper layers, mirroring the formation of expected value representations. Although our approach does not explicitly simulate the recurrent processing, neuromodulator signals, or decision competition between reward representations, our model, by focusing on representational geometry in a high-dimensional sensory network trained under Pavlovian constrains, can be viewed as a complementary study of value learning that emphasizes how both shallow and deep sensory hierarchies together can support the emergence of conditioned representation.

Our associative learning paradigm was grounded in prior human affective studies on classical conditioning. A substantial body of work has leveraged Gabor patches as conditioned stimuli (Song & Keil, 2014; Laurent et al., 2015; Nakashima & Sugita, 2017; Rhodes et al., 2018; Friedl & Keil, 2022). Compared to other types of stimuli, Gabor patches are advantageous in terms of their simple parametric features, allowing precise control over low-level visual properties with minimal confounds. In addition, Gabor filters have been widely used in visual neuroscience to probe early visual cortical processing (Schumacher & Olman, 2010; Bang et al., 2018; Fang et al., 2023), making them a suitable choice for CS+ in our conditioning paradigm for a deep learning model. The selection of two orthogonal orientations serves to maximize the perceptual discriminability and experimental control while enabling concurrent aversive and appetitive conditioning along the same stimulus dimension (orientation). It is important to note that the conditioning effect we report in our study is not limited to a specific stimulus class. Using oriented bars as conditioned stimuli, our model achieved similar conditioning outcomes, in line with successful associative learning from recent rodent studies (Goltstein et al., 2018).

One notable observation is that Gabor patches exhibited neutral valence prior to associative learning (Fig. 3A and 3B). Whereas biological organisms possess the intrinsic ability to recognize the emotional significance of visual input, in the Visual-Valence model, the model's ability to predict the valence of unconditioned stimuli (e.g., IAPS images) is acquired after explicit training. After being trained to recognize the valence of the intrinsically positive or negative images, the model requires features or patterns in the input to drive the valence-related labels. The absence of these features in the Gabor patches will make them to receive a valence assignment around 5 (neutral). During valence learning, the model was assigned positive or negative valence to input images based on features or patterns driving affective responses. The neutrality of low-level visual features defined by the Gabors suggest that features capable of driving affective responses such as valence, likely depend on higher-level visual representations that are semantic- or object-dependent.

In deep learning, reinforcement learning (RL; Sutton & Barto, 2018) provides an established account of how agents adapt their behavior to maximize cumulative reward by reward-prediction-error (RPE) driven learning. As a computational framework, RL has been used to formalize associative learning in human behavior (Gershman, 2015; Law & Gold, 2009). Within this view, operant conditioning is naturally captured as a similar type of learning where policies are shaped by the experienced consequence of choices. By contrast, Pavlovian conditioning is stimulus-centric and typically indexed by autonomic or reflexive responses to conditioned cues. In Pavlovian conditioning, the brain learns to predict the occurrence of an unconditioned stimulus based on the presence of a conditioned stimulus, with the minimization of prediction error. Computationally, this has been modeled as reducing the discrepancy between a conditioned stimulus and an expected unconditioned outcome, closely aligned with temporal difference (TD) learning. Existing studies have leveraged an RL framework (Montague et al., 1996) to show that TD learning can capture key features of classical conditioning. Supervised learning is also closely aligned with Pavlovian conditioning in terms of minimization of prediction error. Our experimental results support the notion of similarity between naturalistic Pavlovian and supervised machine learning by reproducing behavioral and representational hallmarks of empirical observations from human observers (see Appendix D).

The decoding analyses based on neural activation patterns in different layers of the model highlight that associative learning prompts changes in the representations of both the CS+ and the US. The findings regarding the two final fully connected layers in our model, intended to represent anterior emotion-modulated brain regions, support the presence of specific units that are selective for features of both CS+ and US. This is consistent with evidence supporting the role of structures such as the amygdala and orbitofrontal cortex as the locus in which the association between the CS+ and US is represented (Barot et al., 2008; Paton et al., 2006; Romanski et al., 1993). The decoding analysis of classifiers discriminating CS+ and US representations after associative learning (Figs. 5A and 5B) and projection of CS+ and US representations into two-dimensional space (Fig. 5C) suggest that the association of emotional context to neutral cues is established through the co-evolution of the representation patterns of CS+ and US. Specifically, patterns associated with the CS+ and US become more proximate to each other as learning progresses. This suggests that, in a computational architecture, associative learning affects not only the neural representation of the CS+ but also that of the US. This was found in the deeper layers of our neural network, which were built to mimic the role of the amygdaloid complex and other anterior structures linked to emotional processing (Johansen et al., 2010). Based on the behavioral (Baltissen & Boucsein, 1986; Rust, 1976) and fMRI (Dunsmoor et al., 2008) studies of learning related changes of US, an in-depth analysis of how CS+ and US representations influence each other during the learning process is required. Thus, the present study lends support to the hypothesis that a similar process of representational change of CS and US might occur in brain regions where sensory information and emotional outcomes are integrated.

In addition to the findings discussed above, this study contributes to bringing machine learning/AI and computational neuroscience together in a novel way. While our Pavlovian conditioning paradigm falls under the broader category of supervised learning, it diverges significantly from conventional supervised learning in its training and testing methodologies. In our paradigm, supervised learning is first performed by pairing unconditioned stimuli (US) with their associated valence scores (labels). This is followed by supervised training that assigns the compound stimulus consisting of the US and CS+ presently simultaneously, but in different locations, the label of the US. Finally, inference is performed on CS+ only. It is important to note that the present associative learning paradigm differs from associative rule learning (Agrawal et al., 1993), a data mining technique used to identify correlations and co-occurrences between datasets. By integrating associative learning principles from neuroscience into a computational machine learning approach, this study highlights the functional and representational similarities between artificial neurons in deep neural networks (DNN) and biological neurons in the neural system.

Several limitations in this study need to be acknowledged. First, the visual cortex module was kept frozen during the associative learning process, and as such, we were not able to examine the neural plasticity of visual neurons and their influence on the behavioral outcomes in associative learning (McTeague et al., 2015; Miskovic & Keil, 2012; Stolarova et al., 2006). Second, our model lacks sufficient biological realism. Conventional CNNs are pure feedforward models without any temporal processing capability. The absence of recurrence and feedback connections in both visual and emotional networks (Pessoa, 2017; Wyatte et al., 2014) limits the model's capacity to explain the impact of top-down selective attention to cues in associative learning (Mackintosh, 1975; Pearce & Hall, 1980). In addition, our model does not have the ability to account for the effects of neuromodulators such as dopamine (Smith & Torregrossa, 2021; van der Schaaf et al., 2013), which is known to shape and configure aversive and appetitive representation by dopaminergic prediction error. Third, our learning paradigm only has the acquisition phase, whereas the typical associative learning paradigms also contain an extinction phase. This limits us from implementing different phases of conditioning, as well as how the temporal relationship between stimuli is established through learning (Schultz et al., 1997; Sutton, 1988; Sutton & Barto, 1990). Efforts are currently underway to enrich the paradigm so that additional associative learning processes, such as extinction learning, can be tested within the deep learning framework.

**Acknowledgements**

This research was supported by the National Science Foundation grant 1908299 and National Science Foundation grant 2318984, and National Institutes of Mental Health grants R01 MH112558 and R01 MH125615.

## Code Availability

The code for analysis is provided in https://github.com/lab-smile/FearConditioningAI.

## Author Statement*

Author Contributions: SL: Conceptualization, Methodology, Investigation, Visualization, Writing—original draft, Writing—review & editing. AK: Conceptualization, Methodology, Investigation, Supervision, review & editing. RF: Conceptualization, Methodology, Investigation, Supervision, review & editing. MD: Conceptualization, Methodology, Investigation, Supervision, review & editing.